\documentclass[journal, twoside]{IEEEtran}
\usepackage{fancyhdr} 
\usepackage{lipsum}
\usepackage[utf8]{inputenc}
\usepackage[T1]{fontenc}

\usepackage{graphicx} 
\usepackage{subfig}
\usepackage{nccmath}
\usepackage{color}
\usepackage{cite}
\usepackage{acro}
\usepackage{scalerel}
\usepackage{tikz}
\usetikzlibrary{svg.path}
\usepackage{graphicx}
\usepackage[T1]{fontenc}
\usepackage{times}
\usepackage{graphicx}
\usepackage{enumitem}
\usepackage{booktabs}
\usepackage{url}
\usepackage{textcomp}
\usepackage{rotating}
\usepackage{adjustbox}
\usepackage[hidelinks]{hyperref}
\usepackage{subfig}
\usepackage{caption}
\usepackage[utf8]{inputenc}
\usepackage{pdfpages}
\usepackage{CJKutf8}

\definecolor{orcidlogocol}{HTML}{A6CE39}
\tikzset{
  orcidlogo/.pic={
    \fill[orcidlogocol] svg{M256,128c0,70.7-57.3,128-128,128C57.3,256,0,198.7,0,128C0,57.3,57.3,0,128,0C198.7,0,256,57.3,256,128z};
    \fill[white] svg{M86.3,186.2H70.9V79.1h15.4v48.4V186.2z}
                 svg{M108.9,79.1h41.6c39.6,0,57,28.3,57,53.6c0,27.5-21.5,53.6-56.8,53.6h-41.8V79.1z M124.3,172.4h24.5c34.9,0,42.9-26.5,42.9-39.7c0-21.5-13.7-39.7-43.7-39.7h-23.7V172.4z}
                 svg{M88.7,56.8c0,5.5-4.5,10.1-10.1,10.1c-5.6,0-10.1-4.6-10.1-10.1c0-5.6,4.5-10.1,10.1-10.1C84.2,46.7,88.7,51.3,88.7,56.8z};
  }
}

\newcommand\orcidicon[1]{\href{https://orcid.org/#1}{\mbox{\scalerel*{
\begin{tikzpicture}[yscale=-1,transform shape]
\pic{orcidlogo};
\end{tikzpicture}
}{|}}}}

\ifCLASSINFOpdf
\else
\fi
\DeclareAcronym{acm}{
  short = ACM ,
  long  = Association for Computing Machinery ,
  sort  = A ,
}

\usepackage{hyperref} 
\begin{document}

 \setcounter{page}{1}
%
\title{Benchmarking Algorithms for Automatic License Plate Recognition}
%
%
%

\author{Marcel~Del~Castillo~Velarde and Gissel~Velarde
\thanks{Contact Author: Marcel~Del~Castillo~Velarde e-mail:marceldelcastel@gmail.com}
\thanks{Technical Report: November 2021}}

\markboth{Technical Report}{Technical Report}

\maketitle

\begin{abstract}

We evaluated a lightweight Convolutional Neural Network (CNN) called LPRNet \cite{zherzdev2018lprnet} for automatic License Plate Recognition (LPR). We evaluated the algorithm on two datasets, one composed of real license plate images and the other of synthetic license plate images. In addition, we compared its performance against Tesseract \cite{smith2007overview}, an Optical Character Recognition engine. We measured performance based on recognition accuracy and Levenshtein Distance. LPRNet is an end-to-end framework and demonstrated robust performance on both datasets, delivering 90 and 89 percent recognition accuracy on test sets of 1000 real and synthetic license plate images, respectively. Tesseract was not trained using real license plate images and performed well only on the synthetic dataset after pre-processing steps delivering 93 percent recognition accuracy. Finally, Pareto analysis for frequency analysis of misclassified characters allowed us to find in detail which characters were the most conflicting ones according to the percentage of accumulated error. Depending on the region, license plate images possess particular characteristics. Once properly trained, LPRNet can be used to recognize characters from a specific region and dataset. Future work can focus on applying transfer learning to utilize the features learned by LPRNet and fine-tune it given a smaller, newer dataset of license plates.

\end{abstract}

\begin{IEEEkeywords}
License Plate Recognition, Computer Vision, Convolutional Neural Network,  Optical Character Recognition, Pareto analysis

\end{IEEEkeywords}

\IEEEpeerreviewmaketitle


\section{Introduction}

License Plate Recognition (LPR) aims at accurately retrieving a digital string of characters given a License Plate image. The retrieved digitalized License Plate can be stored in a database to perform several tasks automatically for different applications such as Vehicle Traffic Management, Parking Tolls, Digital Surveillance Systems, and Access Control to Buildings. LPR is a solution that allows automatic vehicle control, reducing costs on human labor. Moreover, in some populated countries, the number of vehicles on the streets grows faster than the number of inhabitants. For example, between 1990 and 2015, the vehicle growth rate was 3.5 faster than the growth rate of the Mexican population \cite{IMCO2019}. In Bolivia, the number of vehicles grew by 38 percent between 2014 and 2019 \cite{INE2019}. Therefore, automatic LPR systems are considered to help automize transportation, its management, and security. \\

In this study, we evaluated the performance of a deep network called LPRNet \cite{zherzdev2018lprnet}. We performed tests on two license plates datasets. In addition, we compared its performance against Tesseract \cite{smith2007overview}. While LPRNet was designed to recognize characters from license plates, Tesseract is generally used to recognize characters from documents. Since we had a synthetic dataset available for tests, we selected Tesseract as a baseline for comparison, at least on the synthetic images. We were also interested in testing if LPRNet was able to maintain its performance when trained on a different dataset, given that license plates on each country have unique characteristics. We assumed that a robust system should recognize characters independently of the dataset in use if adequately trained. Besides, we expected LPRNet to be similarly good either on a real and a synthetic dataset. We evaluated recognition accuracy using a test dataset containing $1\,000$ real license plates images \cite{zherzdev2018lprnet} and a subset from a dataset of synthetic license plates images \cite{kaggle}, with $10\,000$ images for training and $1\,000$ images for testing. \\

The contribution of this study is the benchmarking of LPRNet \cite{zherzdev2018lprnet} and Tesseract \cite{smith2007overview} for automatic license plate recognition on two different datasets and the proposal of a framework based on Pareto Analysis \cite{gutierrez2004control} for conflicting characters detection. The article is organized as follows, section \ref{sec:2} presents the problem statement, section \ref{sec:2a} the proposed solution with both analyzed algorithms, section \ref{sec:3} describes the datasets, pre-processing, and evaluation metrics, section \ref{exp} presents the experiments and results. Finally, we conclude in section \ref{sec:5}.

\section{Problem Statement}\label{sec:2}
License Plate Recognition is an application of computer vision for intelligent transportation. Deep learning architectures are currently used to retrieve characters from license plates. Previous models included a segmentation step before recognizing each character, combined with heuristics from the dataset in question \cite{montazzolli2017real}. More recently, end-to-end architectures automatically allow learning properties from raw images for their predictions. End-to-end eliminates the need for intermediate pre-processing steps such as segmentation, thresholding, or the use of heuristics. The advantage of an automatic end-to-end system over a system that relies on heuristics is that by training the model on the dataset in use, the system should be able to discover relevant features on its own. \\

\section{The proposed solution}\label{sec:2a} 
We selected LPRNet for our experiments because its implementation allowed us to train the network from scratch using a selected dataset. Since the alternative dataset we had at disposal for testing was a dataset of synthetic images, we compared LPRNet's performance against Tesseract, which is generally used for Optical Character Recognition from documents. Although Tesseract might not be a natural selection for our application, it allowed us to benchmark LPRNet on the alternative synthetic dataset. If the evaluated algorithms prove to adapt to different datasets, we could expect them to work on a newly collected dataset, given that license plates posses unique characteristics from their particular region.

\subsection{LPRNet}
Convolutional Neural Networks (CNNs) are deep learning models delivering state-of-the-art results in computer vision \cite{alom2019state}. These models simulate neural activation and hierarchical brain processing. In perceptual tasks, CNNs demonstrate to learn relevant representations that are robust to data variability. CNNs learn suitable filters or feature maps on each processing layer, given input images for recognition.   \\

LPRNet is a fast CNN that inputs RGB images of 94x24 pixels and outputs a sequence of characters~\cite{zherzdev2018lprnet}. After all convolutional layers, it follows batch normalization and ReLU activation. LPRNet's architecture is designed to be lightweight and has two main components, a Backbone and Small Basic Blocks. LPRNet uses Connectionist Temporal Classification (CTC) to control the loss for training as end-to-end learning instead of a segmentation process. The Backbone architecture of the network can be seen in Table \ref{tbl:lprnet1}, and the architecture of the Small Basic Blocks can be seen in Table \ref{tbl:lprnet2}. This network was trained with Adam optimizer, batch size 32, gradient noise scale of 0.001, and learning rate of 0.001 that was dropped by a factor of 10 every 100 thousand iterations for a total of 250 thousand iterations \cite{zherzdev2018lprnet}.

\begin{table}[h!]
\begin{center}
\begin{tabular}{ c  p{5cm}}
\toprule
\textbf{Layer Type} & \textbf{Parameters} \\
\midrule
      {Input}
      & 
      {94x24 Pixels RGB image}
      \\  
      {Convolution}
      & 
      {Filters 64, 3x3, stride 1}     
      \\ 
      {Max Pooling}
      & 
      {Filters 64, 3x3, stride 1}   
      \\  
      {Small Basic Block}
      & 
      {Filters 128, 3x3, stride 1} 
      \\ 
      {Max Pooling}
      & 
      {Filters 64, 3x3, stride (2,1)}  
      \\     
      {Small Basic Block}
      & 
      {Filters 256, 3x3, stride 1}       
      \\   
      {Small Basic Block}
      & 
      {Filters 256, 3x3, stride 1}        
      \\ 
      {Max Pooling}
      & 
      {Filters 64, 3x3, stride (2,1)}      
      \\ 
      {Dropout}
      & 
      {0.5 ratio}     
      \\ 
      {Convolution}
      & 
      {Number of class number, 1x13, stride 1}  
      \\ \bottomrule
 
\end{tabular}
\caption[LPRNet]{The LPRNet backbone structure from \cite{zherzdev2018lprnet}.}
\label{tbl:lprnet1}
\end{center}
\end{table}


\begin{table}[h!]
\begin{center}
\begin{tabular}{ c  p{5cm}}
\toprule
\textbf{Layer Type} & \textbf{Parameters} \\
\midrule
      {Input}
      & 
      {$C_{in}$ x H x W features maps}
      \\ 
      {Convolution}
      & 
      {$C_{out}$ /4, 1x1, stride 1}     
      \\ 
      {Convolution}
      & 
      {$C_{out}$ /4, 3x1, stride H=1, padding H=1}    
      \\  
      {Convolution}
      & 
      {$C_{out}$ /4, 1x3, stride W=1, padding W=1}
      \\ 
      {Convolution}
      & 
      {$C_{out}$ 1x1, stride 1} 
      \\ 
      {Output}
      &
      {$C_{out}$ x H x W features maps}
      \\ \bottomrule     
\end{tabular}
\caption[Small Basic Block]{Small Basic Block structure, from \cite{zherzdev2018lprnet}.}
  \label{tbl:lprnet2}
 \end{center}
\end{table}

\subsection{Tesseract}
Tesseract \cite{smith2007overview} is an open source for Optical Character Recognition that expects binary images and outputs a digital string of characters per input. First, Tesseract uses Adaptive Thresholding to binarize images, followed by Connected Component Analysis to find lines and words which are organized as units. Then, if a word is recognized, it aims at finding edges of characters that match their training set. Tesseract was originally trained using $60\,160$ samples of 94 characters, with 20 samples per class in 8 fonts and 4 presentations: normal, italic, bold, and bold-italic  \cite{smith2007overview}.

\subsection{Data Sets}\label{sec:3}
\begin{figure}[h!]
  \centering
  \subfloat[a][Chinese dataset (Real Liscence Plate)]
  {\includegraphics[scale=1]{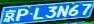}\label{fig:f1}}
  \hfill
  \subfloat[b][Turkey dataset (Synthetic Liscence Plate)]
  {\includegraphics[scale=1]{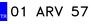}\label{fig:f2}}
  \caption[Examples of images per dataset]{Examples of images per dataset  \cite{zherzdev2018lprnet}, \cite{kaggle}. \ref{fig:f1}, Chinese dataset (real) and \ref{fig:f2}, Turkey dataset (synthetic) .}
\end{figure}

We used two datasets. The first dataset consists of Chinese license plates dataset with $1\,000$ 
RGB real images sized 94x24 pixels, JPG format \cite{zherzdev2018lprnet}. The images are characterized by being taken at different angles and in real situations; see Fig. \ref{fig:f1}. The second dataset consists of a subset from the Synthetic Turkey license plates \cite{kaggle}. We selected the images at random. We used $10\,000$ images for training and $1000$ images for testing. The images are sized 1025x218 pixels in PNG format.  The license plates of this dataset are synthetic images designed in a single frontal view, see Fig. \ref{fig:f2}.

\subsection{Data Pre-Processing}

Turkey dataset images were rescaled to 94x24 pixels to match the dimensions of the Chinese dataset. We used an antialiasing method, which uses a high-quality resolution reduction filter \cite{clark2015pillow}. In addition, PNG images were converted to JPG format.

\begin{figure}[]
\centering
\includegraphics[scale=1]{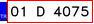}
\caption{Extracting a region of interest and cropping a synthetic license plate from Turkey dataset \cite{kaggle}.}
\label{figc}
\end{figure}

\subsection{Pre-Processing for Tesseract} \label{preT}
Initially, Tesseract delivered poor results on the synthetic dataset. Therefore, we extracted the interest region as seen in Fig. \ref{figc}. Once the interest region was extracted, Tesseract was able to recognize characters accurately from the synthetic Turkey dataset.  However, it was not able to recognize characters from the Chinese dataset. Therefore, we applied image manipulation techniques such as thresholding and binarization before sending the images to Tesseract, but they did not yield significant changes.

\subsection{Evaluation metrics}
We used two metrics for evaluation. First, we evaluated accuracy expressed as:

\begin{equation}
Accuracy=\frac{TP}{TP+TN_{1}+TN_{2}}
\end{equation}

where True Positive (TP) is the correct classification of the license plate, True Negative $TN_{1}$ represents a misclassification when the lengths of the strings between label and prediction are different, and $TN_{2}$ represents a misclassification when the string lengths are equal. This measure only considers the fraction of fully correctly classified license plates over correctly classified and misclassified license plates. However, it could be that a misclassified license plate contains more than one erroneous character. \\

Therefore, we also used the Levenshtein distance between the ground truth labels and the algorithms' outputs. Levenshtein distance measures the smallest number of insertions, deletions, or substitutions needed to transform one sequence of characters to another  \cite{levenshtein1966binary}. See an example in Table \ref{tbl:Levenshtein}.


\begin{table}[!h]
     \begin{tabular}{ c  p{2.3cm}  p{2.3cm}  }
     \toprule
      License Plate & Classification &  Levenshtein Distance \\ 
    \midrule
    \includegraphics[scale=0.95]{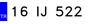}
      & 
      \hspace{10pt}16WW522
      
      & 
      \hspace{20pt}2
   
      \\ 
\includegraphics[scale=0.95]{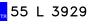}
      & 
      \hspace{10pt}55L3928      
      
      & 
      \hspace{20pt}1
     
      \\ \bottomrule 
 
      \end{tabular}
      \caption[]{Examples of Levenshtein Distance. License plates, classification and corresponding Levenshtein Distance.}
      \label{tbl:Levenshtein}
      \end{table}

\section{Results} \label{exp}
\begin{table}[h]
\begin{center}
\begin{tabular}{c  p{1.18cm} p{1.5cm} p{1.18cm} p{1.3cm} p{1.45cm} }
\toprule
\textbf{Algorithm}&\textbf{Pre-Processing}&\textbf{Dataset}&\textbf{Mean \mbox{Accuracy}}&\textbf{Mean \mbox{Levenshtein} distance}\\
\midrule
 
      {LPRNet}
      &
      {-}
      &
      {Chinese Real images . Train set: $11\,696$ samples. Test set: $1\,000$ samples.}
      & 
      {0.900}
      & 
      {0.047}
   
      \\ 
      {LPRNet}
      & 
      {Resize,RGB to BGR}
      & 
      {Turkey Synthetic images. Train set: $10\,000$ samples. Test set: $1\,000$ samples.}
      & 
      {0.886}
      & 
      {0.030}
          
      \\ 
      {Tesseract}
      & 
      {Resize, Crop}
      & 
      {Turkey Synthetic images. Test set $1\,000$ samples.}
      & 
      {0.933}
      & 
      {0.054}
      \\ \bottomrule 
     
\end{tabular}
\caption[]{Benchmarking LPRNET and Tesseract on real Chinese and synthetic Turkey License Plates datasets. LPRNet's mean accuracy on both datasets is 89 percent. Tesseract achieves 93 percent accuracy on the synthetic dataset.}
\label{tbl:resumen}
\end{center}
\end{table}

We evaluated LPRNet and Tesseract on the synthetic Turkey license plates dataset, as seen in Table \ref{tbl:resumen}. Only LPRNet was evaluated on the real Chinese license plates dataset because we could not train Tesseract on this dataset. We evaluated license plate recognition considering mean accuracy and mean Levenshtein distance. In all cases, we used a test set of $1\,000$ images. We tested the pre-trained LPRNet network. This network was pre-trained on a dataset with $11\,696$ Chinese license plates \cite{zherzdev2018lprnet}. On a test set of $1\,000$ images, LPRNet delivered a mean accuracy of 90 percent and a low mean Levenshtein of 0.047. For the Turkey dataset, images were resized and then converted from RGB to BGR. We re-trained the network with $10\,000$ Turkey license plates, obtaining 88.6 percent mean accuracy and mean Levenshtein of 0.3 on a test set of $1\,000$ images. \\

Tesseract was not trained. On the Chinese test set of $1\,000$ images, Tesseract could not recognize characters. Preliminary tests on the Turkey dataset did not deliver good results either. Therefore, we performed the pre-processing tasks described in section \ref{preT}. We cropped all license plates such that images contained only characters. Then, Tesseract delivered up to 93 percent mean accuracy and a low mean Levenshtein distance. \\

We wondered if Tesseract could deliver similar performance for the Chinese license plates dataset if re-trained on this dataset. However, training Tesseract was out of our possibilities. LPRNet proved robust and appropriate for use on different datasets if properly trained. It is known that for deep learning networks, the more samples for training, the better the results \cite{lecun2015deep}. Our training set for the synthetic dataset was smaller than the training dataset of the real images. It seems possible that with a larger dataset for training, the performance of LPRNet could improve. Given that dataset collection is a costly process, transfer learning could be applied when using LPRNet on a new and much smaller dataset. \\

\begin{figure}
     \begin{tabular}{ c  p{6.5 cm}  p{0 cm}  }
     {\includegraphics[scale=0.32]{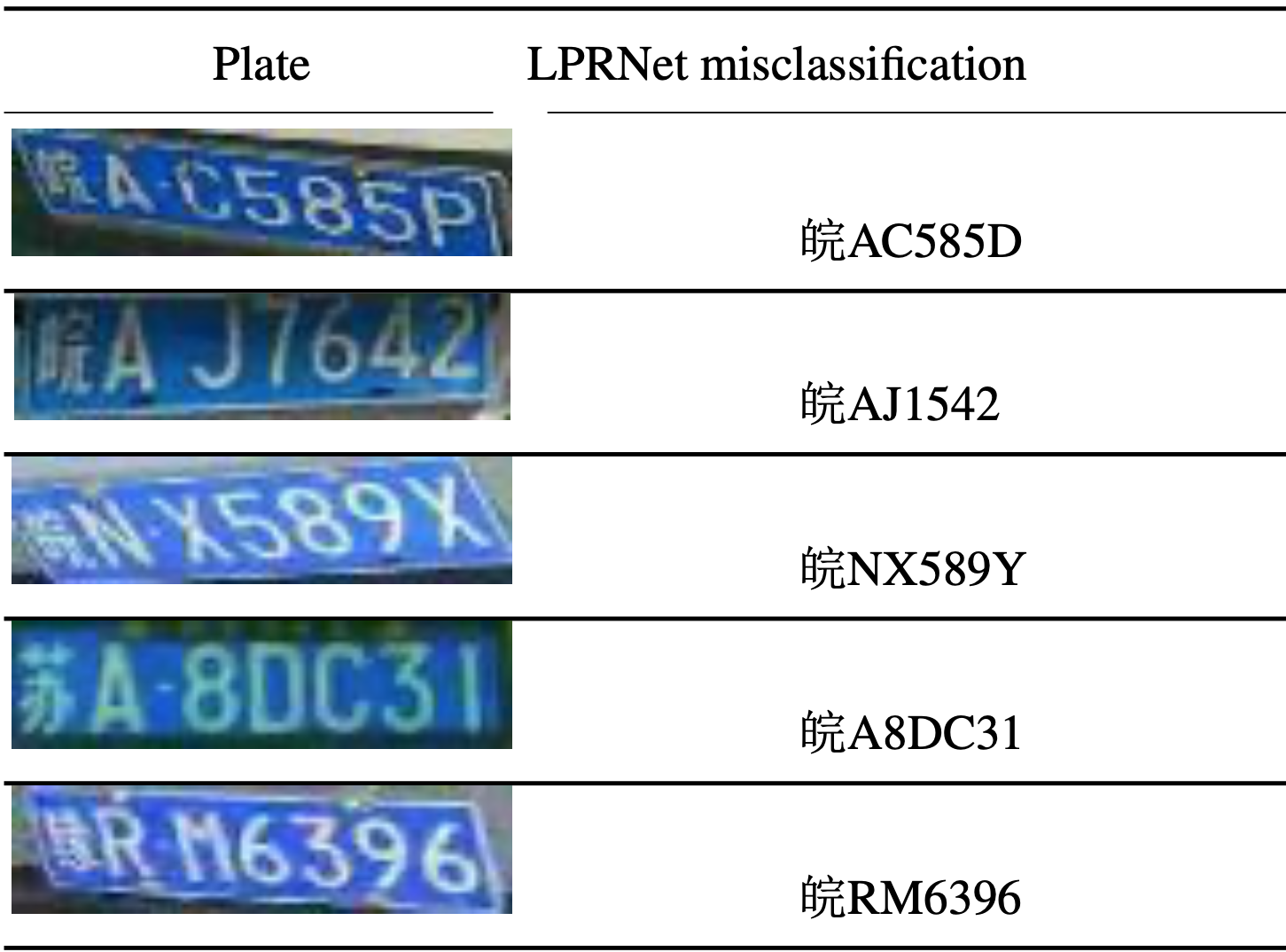}}                    
      \end{tabular}
      \caption{Sample of 5 plates out of 100 misclassified by LPRNet on Chinese dataset. }
      \label{tbl:et1}
\end{figure}

\begin{figure}
     \begin{tabular}{ c p{4.2 cm}  }
     \toprule
      Plate & LPRNet misclassification \\ 
    \cmidrule(r){1-1}\cmidrule(lr){2-2}
     {\includegraphics[scale=0.95]{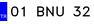}}
      & 
      \hspace{50pt}01BNU35
      \\ \bottomrule
     {\includegraphics[scale=0.95]{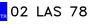}}
      &  
      \hspace{50pt}02L4S78
      \\ \bottomrule
{\includegraphics[scale=0.95]{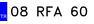}}
      & 
      \hspace{50pt}08RFM60
      \\ \bottomrule
{\includegraphics[scale=0.95]{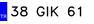}}
      & 
      \hspace{50pt}38GVK61
      \\ \bottomrule
      {\includegraphics[scale=0.95]{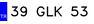}}
      & 
      \hspace{50pt}39GLF53
      \\ \bottomrule                    
      \end{tabular}
      \caption{Sample of 5 plates out of 114 misclassified by LPRNet on Turkish dataset.}
      \label{tbl:et2}
       \end{figure}

\begin{figure}
     \begin{tabular}{ c  p{4.2 cm} }
     \toprule
      Plate & Tesseract misclassification \\ 
    \cmidrule(r){1-1}\cmidrule(lr){2-2}
     {\includegraphics[scale=0.95]{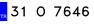}}
      & 
      \hspace{50pt}3107646
      \\ \bottomrule
     {\includegraphics[scale=0.95]{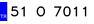}}
      &  
      \hspace{50pt}5107011
      \\ \bottomrule
{\includegraphics[scale=0.95]{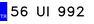}}
      & 
      \hspace{50pt}56Ul992
      \\ \bottomrule
{\includegraphics[scale=0.95]{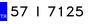}}
      & 
      \hspace{50pt}5717125
      \\ \bottomrule
      {\includegraphics[scale=0.95]{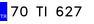}}
      & 
      \hspace{50pt}70Tl627
      \\ \bottomrule                    
      \end{tabular}
      \caption{Sample of 5 plates out of 67 misclassified by Tesseract on Turkish plates.}
      \label{tbl:et3}
       \end{figure}

Fig. \ref{tbl:et1}, Fig. \ref{tbl:et2} and Fig. \ref{tbl:et3} show a subsample of randomly selected plates that were misclassified by each algorithm. Table \ref{tbl:et4} summarizes the number of misclassified license plates on each dataset. In some cases, the length of the output string is the same length as that of the target string. In other cases, the length of the output string has a different length than that of the target string. In the next section, we evaluated misclassified characters of outputs with the same length. 
\begin{table}[!h]
     \begin{tabular}{ c  p{1.2 cm}  p{1.2 cm} p{1.2 cm} p{1.4cm} }
     \toprule
      Algorithm & Test set & Same length & Diferent length & Total Misclassified\\ 
    \midrule
      {LPRNet}
      &
      {Chinese}
      &
      {40}
      &
      {60}
      &
      {100}
      \\ 
      {LPRNet}
      &
      {Turkish}
      &
      {26}
      &
      {88}
      &
      {114}
      \\ 
      {Tesseract}
      &
      {Turkish}
      &
      {49}
      &
      {18}
      &
      {67}

      \\ \bottomrule                  
      \end{tabular}
      \caption{Number of misclassified license plates depending
on the length.}
      \label{tbl:et4}
      \end{table}

\subsection{Pareto analysis} \label{sec:4}
We used Pareto analysis for both algorithms (Tesseract and LPRNet) to identify misclassified license plates, where the predicted output had the same length as the target string. Pareto analysis obtains a percentage of frequency of misclassified characters and arranges these frequencies in descent order, from most conflicting to least  \cite{gutierrez2004control}. This analysis helped us identify which characters were the most conflicting characters. We observed:
\begin{itemize}
\item False Positive Characters, and
\item False Negative Characters.
\end{itemize}

\begin{figure}[]
 \begin{center}
 \includegraphics[scale=.46]{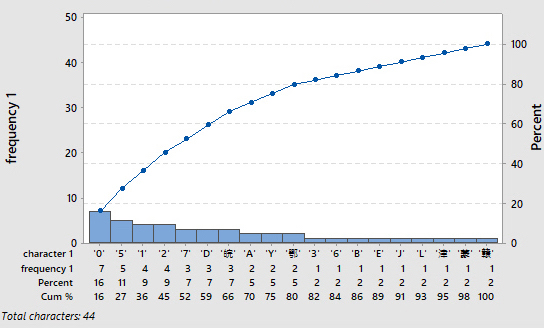}
  \caption{False Positive Characters. Pareto chart for LPRNet. Set: 40 of 1000 Chinese license plates misclassified (same length).  }
  \label{fig:pareto1}
 \end{center}
\end{figure}

\begin{figure}[]
 \begin{center}
 \includegraphics[scale=.46]{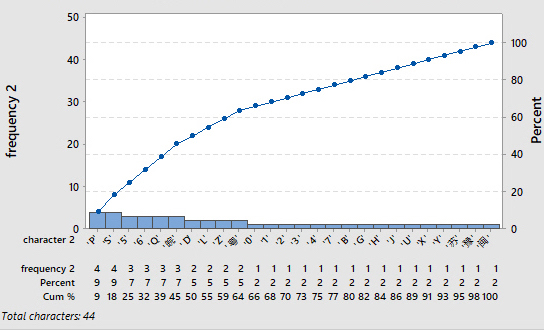}
  \caption{False Negative Characters. Pareto chart for LPRNet. Set: 40 of 1000 Chinese license plates misclassified (same length). }
  \label{fig:pareto2}
 \end{center}
\end{figure}

\begin{figure}[]
 \begin{center}
 \includegraphics[scale=.46]{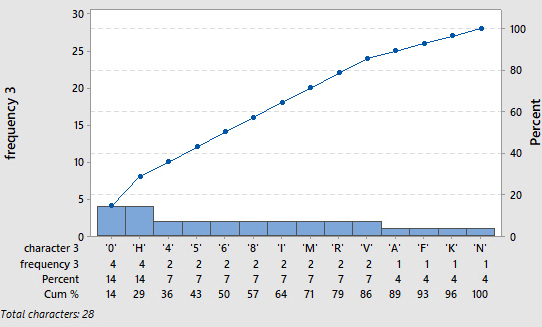}
  \caption{False Positive Characters. Pareto chart for LPRNet. Set: 26 of 1000 synthetic Turkey license plates misclassified (same length). }
  \label{fig:pareto3}
 \end{center}
\end{figure}

\begin{figure}[]
 \begin{center}
 \includegraphics[scale=.46]{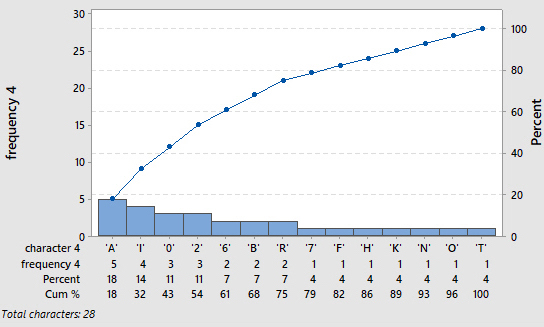}
  \caption{False Negative Characters. Pareto chart for LPRNet. Set: 26 of 1000 synthetic Turkey license plates misclassified (same length). }
  \label{fig:pareto4}
 \end{center}
\end{figure}

\begin{figure}[]
 \begin{center}
 \includegraphics[scale=.46]{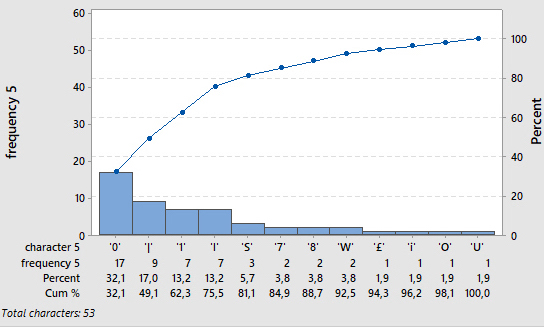}
  \caption{False Positive Characters. Pareto chart for Tesseract. Set: 49 of 1000 synthetic Turkey license plates misclassified (same length). }
  \label{fig:pareto5}
 \end{center}
\end{figure}

\begin{figure}[]
 \begin{center}
 \includegraphics[scale=.46]{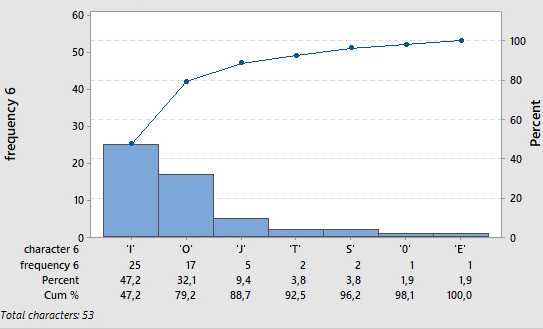}
  \caption{False Negative Characters. Pareto chart for Tesseract. Set: 49 of 1000 synthetic Turkey license plates misclassified (same length). }
  \label{fig:pareto6}
 \end{center}
\end{figure}

False Positive Characters are predicted characters misclassified by the algorithm; that is, the algorithm predicted a `P' but the target was not a `P'. False Negative Characters are target characters misclassified by the algorithm or characters that should have been predicted, but were not recognized. For example, the target character was an `A', but the algorithm predicted another character.  Therefore, for each experiment, we created two Pareto charts identifying the most conflicting characters, the percentage of error for each character and its accumulated error, see Fig. \ref{fig:pareto1} to Fig. \ref{fig:pareto6}.\\

Since we could not train Tesseract on Chinese dataset, we performed a Pareto analysis only for LPRNet on the Chinese dataset, row 1 in Table \ref{tbl:resumen}. The analysis was performed on a set of 40 out of 1000 license plates misclassified such that the output string of LPRNet had the same length as the target license plate string. Erroneous outputs with strings of different lengths were not considered in this analysis. Pareto charts for False Positive Characters and False Negative Characters are shown in Fig. \ref{fig:pareto1} and Fig. \ref{fig:pareto2}, respectively. We observed that `0' and `5' are the most frequent False Positive Characters, while `P', `S', and `5' are the most frequent False Negative Characters. In the synthetic Turkey dataset, we analyzed 26 out of 1000 license plates with the same length as the target strings. Similarly, LPRNet outputs `0' as a frequent False Positive Character as well as `H'. The three most frequent False Negative Characters were `A', `I' and `0', see Fig. \ref{fig:pareto3} and Fig. \ref{fig:pareto4}.\\

Finally, Fig. \ref{fig:pareto5} and Fig. \ref{fig:pareto6} are Pareto charts for Tesseract on a set of 49 misclassified license plates (same length), row 3 in Table \ref{tbl:resumen}. Like for LPRNet, `0' is the most frequent False Positive Character, with 17 occurrences, while `I' and `O' are the most frequent False Negative Characters. In general, distinguishing `0' from `O' seems to be the most challenging pair of characters. 

\section{Conclusion}\label{sec:5}
LPRNet demonstrated robustness in recognizing characters from both license plate datasets, either from real images or synthetic images, given that the network is properly trained with the corresponding dataset. Tesseract was not trained on real license plate images and performed well on the synthetic dataset only. We wondered if Tesseract could also perform well on the real images from the Chinese dataset, if adequately trained on it. However, training Tesseract was out of our reach. Pareto analysis helped us identify the misclassified character frequencies automatically. For both algorithms, distinguishing `0' from `O' proved to be difficult. Without Pareto analysis, it would be time consuming to discover which are the conflicting characters. LPRNet, a lightweight deep learning network, demonstrated to learn relevant filters for a given dataset. We observe that LPRNet's reported performance can be achieved with at least $10\,000$ license plates images. Collecting this amount of sample images could be a long and costly process. Future work could investigate applying transfer learning to utilize the features learned by LPRNet, and fine-tune the network on a new and smaller dataset from a particular region. The application of smart technologies for automatic license plate recognition constitute a step towards better transportation, its management, and security. \\

%
%
%



%


\bibliographystyle{ieeetr}  

\bibliography{thebibliography}
\ifCLASSOPTIONcaptionsoff
  \newpage
\fi

\end{document}